%% file: root.tex
\pdfoutput=1
\documentclass[letterpaper,10pt,conference]{ieeeconf}

\IEEEoverridecommandlockouts
\usepackage[T1]{fontenc}
\usepackage{graphicx}
\usepackage{amsmath,amssymb}
\usepackage{booktabs}
\usepackage{array}
\usepackage{cite}
\usepackage{url}
\usepackage{acronym}
\input{acronyms}

\urldef{\projectwebsite}\url{https://capability-aware.github.io/}
\newcommand{\method}{TempoLoco}

\title{\LARGE \bf
Beyond Fixed Visual History: Learning Motion-Adaptive Temporal Receptive Fields for Complex-Terrain Quadruped Locomotion
}

\author{
Hao Zhang$^{1}$,
Yuping Yan$^{1}$,
Yitian Hong$^{2}$,
Luying Feng$^{1}$,\\
Tianhong Wang$^{1}$,
Fengqi Dai$^{1}$,
Wei Chen$^{1}$,
Yaochu Jin$^{1}$%
\thanks{$^{1}$Westlake University, Hangzhou, China.}%
\thanks{$^{2}$East China University of Science and Technology,
Shanghai, China.}%
\thanks{Project website:
\projectwebsite}%
}

\begin{document}

\maketitle
\thispagestyle{empty}
\pagestyle{empty}


\begin{abstract}
Versatile quadruped locomotion depends on a tight coupling between what the robot perceives and what its body can execute.  Yet existing visual controllers typically organize depth history using fixed temporal patterns, overlooking that the same temporal span corresponds to different spatial ranges as the robot's speed, posture, and gait phase change. Consequently, the same visual history may be insufficient in one motion state but outdated or misaligned in another. To address this limitation, we propose TempoLoco, a body-aware dynamic visual perception framework that organizes visual memory according to the robot's current locomotion state. TempoLoco learns multiple complementary temporal views from a long depth buffer, encodes them within a shared feature space, and dynamically aggregates them using proprioceptive history and multi-scale terrain evidence. This enables a single policy to align visual anticipation with the robot's current motion capabilities across complex terrains, including stairs, platforms, gaps, discrete obstacles, rough ground, and slopes. From the experiment analysis, TempoLoco achieves a 78.4\% success rate, compared with 58.2\% for the strongest fixed-history baseline and 62.1\% for uniform multi-scale fusion. In separate capability evaluations, it completes 41/45 simulated trials across nine terrain types and 23/25 real-world trials on a Unitree B2 with a fixed arm, demonstrating robust and deployable locomotion through body-aware dynamic visual perception.
\end{abstract}


\section{Introduction}
\label{sec:introduction}

\begin{figure*}[t]
    \centering
    \includegraphics[width=\textwidth]{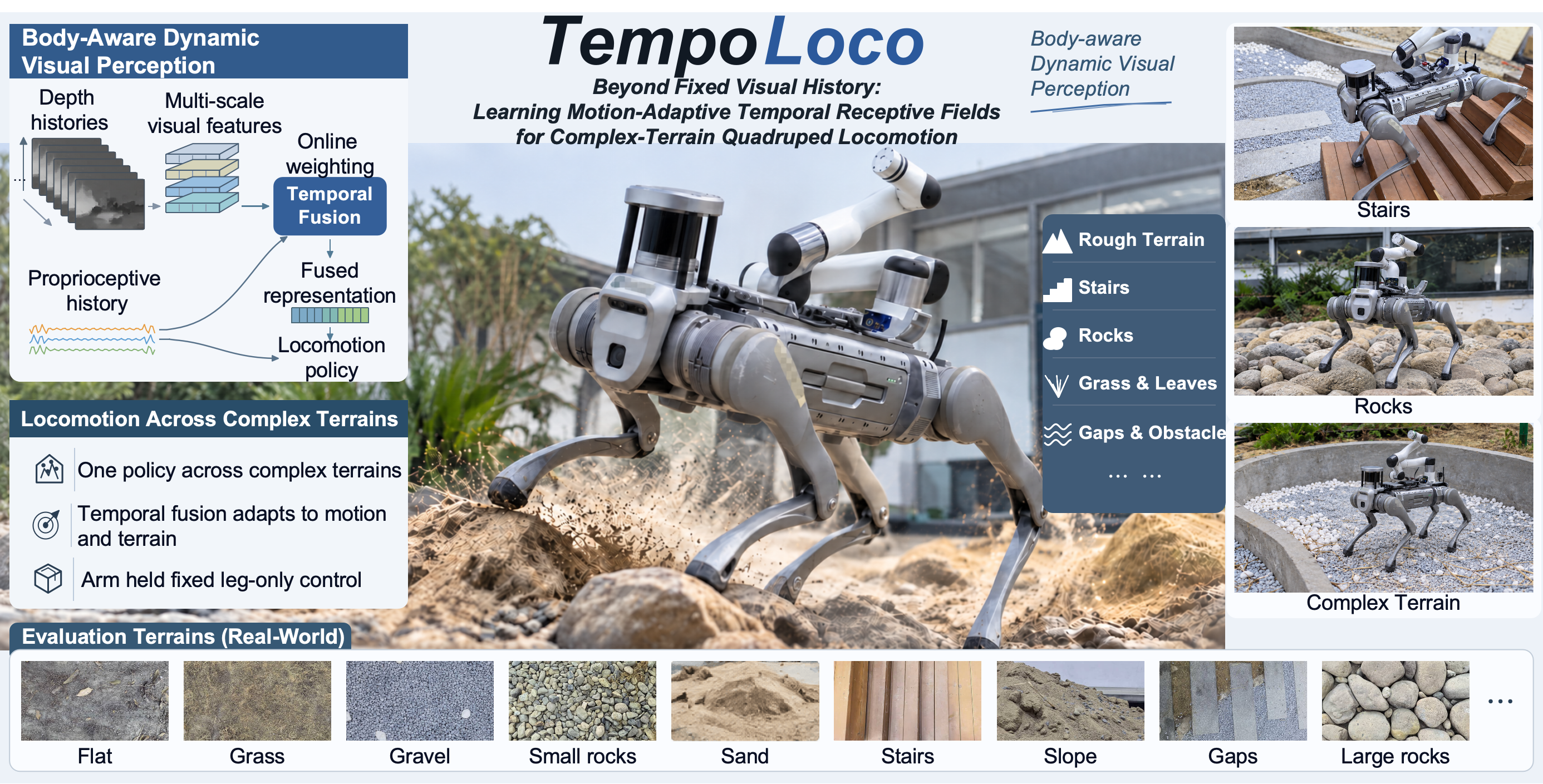}
\caption{Overview of \method{}. Globally learned temporal strides construct complementary depth histories, which are dynamically aggregated by a body-state-conditioned gate using proprioceptive history and multi-scale terrain evidence. The resulting visual representation drives a single leg-control policy across diverse complex terrains. The strides remain fixed after training, whereas the fusion weights adapt online at every policy step. The mounted arm is fixed and excluded from the locomotion action space.
}
    \label{fig:overview}
\end{figure*}

Versatile quadruped locomotion over complex terrain requires a robot not only to perceive terrain geometry but also to interpret it in relation to its current motion capabilities~\cite{hwangbo2019agile,tan2018sim2real,ha2025survey}. As speed, posture, and gait phase evolve, both the robot's motion capability and the relevance of previously observed terrain change continuously~\cite{kumar2021rma,dreamwaq2023,margolis2023walk}. However, existing approaches have primarily focused on how visual and proprioceptive features are fused, while the temporal organization of visual information is usually predetermined and remains unchanged throughout locomotion \cite{miki2022perceptive,hoeller2024anymal,agarwal2023egocentric,kareer2023vinl}. This creates a fundamental mismatch between a dynamically evolving body and a statically organized visual memory.

Depth-history construction determines the temporal evidence available to the controller and can be broadly categorized into dense and sparse sampling strategies~\cite{zhu2026hiking}. Under a fixed frame budget, dense sampling preserves recent geometric changes and remains closely aligned with the robot's current viewpoint, but covers only a short approach history~\cite{zhang2026transpercept}. Sparse sampling extends farther into the past and can retain terrain that has already left the current field of view, but increasingly incorporates observations captured from different body positions, orientations, and gait phases~\cite{zhu2026hiking,miki2022perceptive}. Moreover, the same temporal span corresponds to different traveled distances as locomotion speed changes. Visual history is therefore not merely a sequence of past observations, but a \emph{body-relative perceptual field} whose physical meaning changes with the robot's motion state. A visual history that is useful in one state may consequently become insufficient, redundant, or misaligned in another.

To exploit a broader visual history, existing perceptive locomotion methods typically select observations using predefined temporal patterns, such as fixed frame stacking, sliding windows, or fixed-stride sampling, and then encode or reweight them using recurrent networks, attention mechanisms, or proprioception-conditioned feature modulation~\cite{miki2022perceptive,yang2021locotransformer,zhang2026transpercept,liu2026real,agarwal2023egocentric,zhu2026hiking}. Although these mechanisms improve the representation of the selected observations, the available terrain evidence remains constrained by the original history construction. The central problem is therefore not only \emph{how to encode visual history}, but also \emph{which temporal views should be made available and how their relevance should change with the robot's current motion state}.


We therefore propose \method{} (Fig.~\ref{fig:overview}), a motion-adaptive temporal perception framework that separates \emph{global temporal-scale learning} from \emph{state-conditioned fusion}. \method{} learns a bank of continuous temporal strides and uses differentiable sampling to construct multiple histories from a longer depth buffer. A shared visual encoder maps these histories into a common feature space, after which a temporal gate combines them according to proprioceptive history and multi-scale visual evidence. The candidate strides are global learned parameters, while their fusion weights adapt at each policy step. This enables temporal adaptation within a single locomotion policy without switching among independently trained controllers or changing the downstream action interface. Our contributions are threefold:


\begin{itemize}

    \item We formulate the terrain- and motion-dependent nature of visual temporal context in quadruped locomotion. Fixed Current, Dense, and Sparse histories exhibit performance reversals and complementary successes, showing that no single temporal pattern is uniformly effective.

    \item We introduce \method{}, which learns complementary continuous temporal views and dynamically aggregates them using proprioceptive history and multi-scale terrain evidence. By separating global scale discovery from state-conditioned scale utilization, \method{} realizes body-aware dynamic visual perception within a single locomotion policy.

    \item We conduct extensive evaluations with matched controls across training seeds, terrain difficulty, simulation, and real-world deployment. \method{} achieves $78.4\pm2.3\%$ success over the full ten-level curriculum, improves mean worst-terrain success from $32.1\%$ to $54.5\%$, and completes 41/45 simulation trials and 23/25 physical trials on a Unitree B2.

\end{itemize}


\section{Related Work}
\label{sec:related_work}

\subsection{Perceptive Legged Locomotion}

Perceptive quadruped locomotion requires coupling terrain observations with the robot's evolving body state so that control decisions remain compatible with both upcoming geometry and current motion capabilities. Building on robust proprioceptive sim-to-real control and scalable
learning-based locomotion~\cite{tan2018sim2real,kumar2021rma,
lee2020challenging,rudin2022walk,margolis2023walk,ha2025survey}, recent methods incorporate exteroception through recurrent belief
representations~\cite{miki2022perceptive}, cross-modal attention~\cite{yang2021locotransformer}, and visual policies for agile obstacle traversal~\cite{hoeller2024anymal,
cheng2024extreme,agarwal2023egocentric,kareer2023vinl,seo2023steering}. More recent systems, including Hiking in the Wild~\cite{zhu2026hiking} and MGDP~\cite{dong2026mgdp,grandia2023perceptive}, further extend perceptive locomotion to diverse and challenging terrains.

Collectively, these studies establish the importance of visual--proprioceptive coupling for complex-terrain locomotion. Their primary focus, however, lies in control learning, cross-modal feature integration, or overall locomotion capability, while the temporal organization of visual history is typically predetermined. Our work therefore investigates a complementary question: \emph{how should depth
history be constructed and utilized so that visual perception remains aligned with the robot's evolving motion state?}

\subsection{Temporal Modeling for Perceptive Locomotion}

Temporal information is commonly incorporated into perceptive
locomotion through frame stacking, recurrent memory, or sequence
models~\cite{miki2022perceptive,agarwal2023egocentric,
zhang2026transpercept,zhu2026hiking}.
Recurrent belief representations retain information across
control steps~\cite{miki2022perceptive}, and MGDP uses
recurrent processing to construct history-aware terrain
features~\cite{dong2026mgdp}. Trans-Percept applies causal
attention over a sliding multimodal
history~\cite{zhang2026transpercept}, while Hiking in the Wild
constructs temporal depth inputs using a prescribed strided
sampling pattern~\cite{zhu2026hiking}.

These approaches improve how temporal observations are
retained, encoded, or aggregated. Nevertheless, they generally
operate on visual histories whose window length and sampling
schedule are specified before encoding. Consequently, the
temporal model can emphasize or suppress observations already
contained in the input history, but it cannot access evidence
excluded by its predefined construction.

\subsection{State-Conditioned Visual Perception}

Robot state has also been used to adapt visual processing.
LocoTransformer~\cite{yang2021locotransformer} performs
state-dependent cross-modal attention between proprioceptive
and visual features. REAL~\cite{liu2026real} uses
proprioception-conditioned FiLM modulation together with a
Mamba-based temporal backbone, enabling the visual
representation to respond to the robot's current dynamics.
These methods demonstrate the value of interpreting visual
information in relation to body motion. However, their temporal
input support is typically determined before feature encoding;
state conditioning primarily acts on the representation of the
available observations.


\section{Method}
\label{sec:method}

\begin{figure*}[t]
    \centering
    \includegraphics[width=\textwidth]{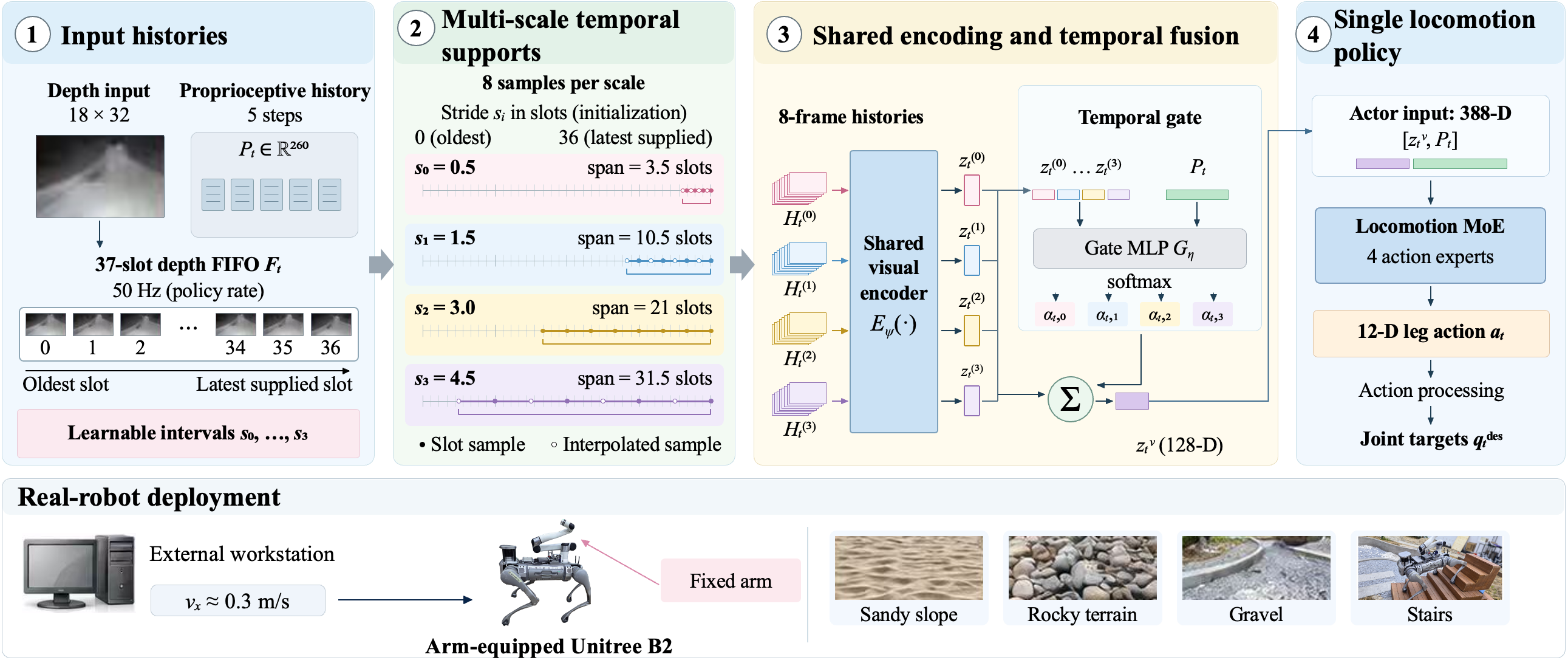}
\caption{Detailed architecture of \method{} (actor path shown). At each 50-Hz policy step, a 37-slot depth \acs{FIFO} and a five-step, 260-dimensional proprioceptive history are provided to the temporal-perception module. Four globally learned continuous strides construct complementary eight-frame histories; the displayed stride values are the initialization, and filled and hollow markers denote integer-slot and interpolated samples, respectively. A scale-shared visual encoder maps each history to a 128-dimensional feature, and a body-state-conditioned temporal gate computes online softmax fusion weights from proprioceptive history and all scale-specific visual features. The fused 128-dimensional visual representation is concatenated with proprioception to form the 388-dimensional actor input, which is processed by the four-expert locomotion \acs{MoE} and the unchanged action-processing pipeline to produce 12 leg-joint targets. The actor and critic use independent temporal modules, and the critic appends 33 privileged dimensions only after temporal fusion. The lower panel shows deployment on the arm-equipped Unitree B2, with the arm fixed outside the locomotion action space.}
    \label{fig:architecture}
\end{figure*}

\subsection{Body-Aware Dynamic Visual Perception Overview}

\method{} realizes body-aware dynamic visual perception by
separating visual-memory construction into two complementary
processes: \emph{global temporal-view discovery} and
\emph{state-conditioned temporal-view utilization}. The former
learns a compact set of temporal views that capture terrain
evidence over different look-back ranges, while the latter
determines how these views should contribute under the robot's
current motion state. This decomposition allows the visual
representation to evolve with the robot's speed, posture, and
gait phase without changing the downstream locomotion policy
or action interface.

At policy step $t$, the controller receives a five-step
proprioceptive history
\begin{equation}
    \mathbf{P}_t =
    [\mathbf{p}_{t-4},\ldots,\mathbf{p}_t]
    \in \mathbb{R}^{260},
\end{equation}
where each $\mathbf{p}_t\in\mathbb{R}^{52}$ contains estimated
body-frame linear velocity, angular velocity, projected gravity,
velocity commands, joint positions and velocities, the previous
clipped action, and gait phase. The body-frame linear velocity
is provided by a separately supervised $240\!\rightarrow\!3$
estimator that is optimized independently of \ac{PPO}.

The visual input is maintained as an oldest-to-newest
policy-time depth memory
\begin{equation}
    \mathbf{F}_t =
    [\mathbf{D}_{t-T+1},\ldots,\mathbf{D}_t]
    \in \mathbb{R}^{T\times H\times W},
\end{equation}
where $T=37$, $H=18$, and $W=32$. Here, $\mathbf{D}_t$
denotes the latest depth observation supplied to the policy;
adjacent \ac{FIFO} slots need not correspond to distinct physical
camera captures.

Rather than encoding the complete buffer directly, \method{}
factorizes body-aware visual perception as
\begin{equation}
\begin{aligned}
    \{\mathbf{H}^{(i)}_t\}_{i=0}^{M-1}
        &= \mathcal{S}_{\boldsymbol{\theta}}(\mathbf{F}_t),\\
    \mathbf{z}^{(i)}_t
        &= E_{\psi}(\mathbf{H}^{(i)}_t),\\
    \boldsymbol{\alpha}_t
        &= G_{\eta}
        \left(
        \mathbf{P}_t,
        \mathbf{z}^{(0)}_t,\ldots,
        \mathbf{z}^{(M-1)}_t
        \right),\\
    \mathbf{z}^{v}_t
        &= \sum_{i=0}^{M-1}
        \alpha_{t,i}\mathbf{z}^{(i)}_t,
\end{aligned}
\end{equation}
where $\mathcal{S}_{\boldsymbol{\theta}}$ constructs
complementary temporal views, $E_{\psi}$ is a scale-shared
visual encoder, and $G_{\eta}$ is a body-conditioned temporal
gate. The resulting $\mathbf{z}^{v}_t\in\mathbb{R}^{128}$
retains the original visual interface of the locomotion
backbone.

The actor and critic instantiate independent stride banks,
visual encoders, and temporal gates. The critic additionally
receives privileged information only after temporal fusion;
therefore, temporal-view utilization depends exclusively on
observations available to the deployed policy.

\subsection{Learnable Temporal Receptive Fields}

A fixed visual history imposes the same balance between recent
geometric detail and longer-range approach information on all
terrain and motion states. To provide complementary temporal
contexts, each actor or critic branch learns $M=4$ continuous
temporal strides:
\begin{equation}
    s_i = 5\sigma(\theta_i),
    \qquad i=0,\ldots,M-1,
\label{eq:stride}
\end{equation}
where $\sigma(\cdot)$ is the sigmoid function. The strides are
initialized to $[0.5,\ 1.5,\ 3.0,\ 4.5]$.

The stride bank is globally learned during training. Its values
are shared across observations rather than predicted separately
at every policy step. Actor and critic use independent stride
banks, no ordering constraint is imposed, and no branch is fixed
to a current-only history.

For each stride $s_i$, the method constructs a temporal view
containing $K=8$ depth observations. The continuous location of
the $k$-th observation is
\begin{equation}
    r_{i,k}
    =
    \operatorname{clip}
    \left(
        T-1-s_i(K-1-k),
        0,
        T-1
    \right),
\label{eq:sample_position}
\end{equation}
Every temporal view is anchored at the latest supplied depth
slot. A smaller stride densely samples recent geometry, whereas
a larger stride covers a longer portion of the terrain approach
under the same eight-frame budget.

Since $r_{i,k}$ is generally non-integer, the corresponding
observation is obtained through linear interpolation. Let
\begin{equation}
\begin{aligned}
    l_{i,k} &= \lfloor r_{i,k}\rfloor,\\
    u_{i,k} &= \min(l_{i,k}+1,T-1),\\
    \beta_{i,k} &= r_{i,k}-l_{i,k}.
\end{aligned}
\end{equation}
The interpolated depth observation is
\begin{equation}
    \widetilde{\mathbf{D}}_{i,k}
    =
    (1-\beta_{i,k})\mathbf{F}_t[l_{i,k}]
    +
    \beta_{i,k}\mathbf{F}_t[u_{i,k}],
\label{eq:interpolation}
\end{equation}
and the resulting oldest-to-newest temporal view is
\begin{equation}
    \mathbf{H}^{(i)}_t
    =
    [
        \widetilde{\mathbf{D}}_{i,0},
        \ldots,
        \widetilde{\mathbf{D}}_{i,K-1}
    ]
    \in
    \mathbb{R}^{8\times18\times32}.
\end{equation}

For a fixed \ac{FIFO}, linear interpolation is continuous and
piecewise differentiable with respect to $s_i$. Gradients from
the policy and value objectives can therefore optimize the
temporal sampling locations through the interpolation
coefficients. This provides end-to-end temporal receptive field
learning without discrete frame selection or an additional
sampling policy.

The sampling operation acts in \ac{FIFO}-index space. It does not
perform ego-motion compensation, geometric reprojection, or
interpolation over physical camera timestamps. Its purpose is
to expose the controller to complementary learnable views of
the supplied visual memory.

\subsection{State-Conditioned Dynamic Fusion}

Each temporal view is mapped into a common visual feature space:
\begin{equation}
    \mathbf{z}^{(i)}_t
    =
    E_{\psi}(\mathbf{H}^{(i)}_t)
    \in \mathbb{R}^{128}.
\end{equation}
Within each actor or critic branch, all temporal views share one
visual encoder. The encoder applies a $3\times3$ convolution
from eight input channels to four output channels, followed by
\ac{ReLU}, $2\times2$ max pooling, and fully connected layers $576\rightarrow256\rightarrow256\rightarrow128$.
Parameter sharing places all temporal views in a common feature
space while limiting parameter growth. The actor and critic
retain separate visual encoders.

The utility of a temporal view depends on both the robot's
current motion state and the terrain evidence contained in that
view. We therefore construct
\begin{equation}
    \mathbf{c}_t
    =
    [
        \mathbf{P}_t,
        \mathbf{z}^{(0)}_t,
        \ldots,
        \mathbf{z}^{(M-1)}_t
    ]
    \in \mathbb{R}^{772}
\end{equation}
and compute
\begin{equation}
\begin{aligned}
    \boldsymbol{\alpha}_t
        &=
        \operatorname{softmax}
        \left(G_{\eta}(\mathbf{c}_t)\right),\\
    \mathbf{z}^{v}_t
        &=
        \sum_{i=0}^{M-1}
        \alpha_{t,i}\mathbf{z}^{(i)}_t,
\end{aligned}
\label{eq:temporal_fusion}
\end{equation}
where $G_{\eta}$ is a
$772\rightarrow64\rightarrow4$ temporal gate,
$\alpha_{t,i}\geq0$, and
$\sum_i\alpha_{t,i}=1$. The output layer of the gate is
initialized to zero, yielding uniform initial fusion weights.

The stride bank and temporal gate address different aspects of
dynamic perception. The stride bank discovers a stable set of
candidate temporal receptive fields during training, whereas
the gate determines their relevance from the current
visual--proprioceptive context. The learned strides remain fixed
at deployment, while $\boldsymbol{\alpha}_t$ is recomputed at
every policy step. Thus, online motion adaptation arises from
state-conditioned utilization of complementary temporal views,
rather than from predicting new stride values at each step.

For the actor, the fused visual representation is concatenated
with proprioception:
\begin{equation}
    \mathbf{x}^{a}_t
    =
    [\mathbf{P}_t,\mathbf{z}^{v}_t]
    \in \mathbb{R}^{388}.
\end{equation}
The critic independently constructs its fused visual feature
and appends $33$ privileged dimensions afterward:
\begin{equation}
    \mathbf{x}^{c}_t
    =
    [
        \mathbf{P}_t,
        \mathbf{z}^{v,c}_t,
        \boldsymbol{\xi}_t
    ]
    \in \mathbb{R}^{421}.
\end{equation}
The privileged variables include randomized control gains,
base pitch and height, clean linear velocity, and foot-contact
indicators. They are excluded from the temporal gate, ensuring
that dynamic temporal fusion relies only on deployable
observations.

\subsection{Joint Policy Learning and Stabilization}

The proposed temporal module is inserted before the existing
four-expert locomotion mixture-of-experts backbone. The
temporal gate in Eq.~\eqref{eq:temporal_fusion} determines how
visual histories are fused, whereas the locomotion \ac{MoE} gate
determines how action experts are combined. These are two
distinct mechanisms.

Each actor expert is a \ac{MLP} with \ac{ELU} activations and architecture $388\rightarrow256\rightarrow128\rightarrow64\rightarrow12$, while the action-expert gate has architecture $388\rightarrow64\rightarrow4$. The weighted sum of the expert outputs defines the mean of a diagonal Gaussian action distribution. The mean is smoothly bounded by
$
    \boldsymbol{\mu}_t
    =
    3\tanh
    \left(
        \frac{\widetilde{\boldsymbol{\mu}}_t}{3}
    \right)$.
Training samples from the Gaussian policy, while deterministic
evaluation and hardware deployment use
$\boldsymbol{\mu}_t$. \ac{PPO} jointly optimizes the locomotion
policy, the temporal strides, the shared visual encoders, and
the temporal gates.

Jointly learning the temporal receptive fields and their
state-dependent fusion may cause an early dominant branch to
suppress the learning of the remaining views. We therefore use
three complementary stabilization mechanisms.

\paragraph{Uniform-fusion warm-up.}
During the first $3{,}000$ \ac{PPO} iterations, the temporal gate is
bypassed and all temporal views are fused uniformly by $\alpha_{t,i}=\frac{1}{M}$. The temporal gate receives no gradient during this stage, while
the stride bank, visual encoder, and locomotion policy remain
trainable. State-conditioned fusion is enabled after the
multi-view representation has been initialized.

\paragraph{Temporal-support diversity.}
To discourage different strides from converging to the same
temporal support, we use
\begin{equation}
    \mathcal{R}_{\mathrm{div}}
    =
    \sum_{i<j}
    \exp
    \left(
        -\frac{|s_i-s_j|}{\tau_s}
    \right),
\label{eq:diversity}
\end{equation}
where $\tau_s=0.5$ and
$\lambda_{\mathrm{div}}=10^{-4}$. This term encourages
separation among the learned temporal supports without
enforcing an ordering or prescribing their final values.

\paragraph{Branch-utilization balancing.}
After activating the temporal gate, persistent underuse of a
temporal view is discouraged through
\begin{equation}
    \mathcal{R}_{\mathrm{bal}}
    =
    \sum_{i=0}^{M-1}
    \left[
        \rho-
        \frac{1}{B}
        \sum_{n=1}^{B}\alpha_{n,i}
    \right]_{+}^{2},
\label{eq:balance}
\end{equation}
where $B$ is the \ac{PPO} minibatch size, $\rho=0.10$, and
$\lambda_{\mathrm{bal}}=0.01$. The regularizer acts on average
soft fusion weights rather than hard assignments, allowing
nonuniform state-dependent preferences while reducing
persistent branch starvation.

In implementation, the weighted gradients associated with
Eqs.~\eqref{eq:diversity} and~\eqref{eq:balance} are applied
to the stride logits and temporal-gate logits, respectively,
before global gradient-norm clipping and the Adam update.
These mechanisms affect training only and do not alter the
forward computation during evaluation or deployment.

The downstream action space and low-level control interface are
unchanged across all methods. The policy outputs twelve leg
actions, while the mounted arm remains fixed outside the
locomotion action space. Action filtering, joint-target
construction, \ac{PD} tracking, and hardware control rates are
described in Sec.~IV-A.

\begin{table*}[!t]
    \centering
    \caption{Fixed-history diagnosis: 9,000 episodes. Values are percentages; overall results are training-seed mean$\pm$SD.}
    \label{tab:fixed_diagnosis}
    \setlength{\tabcolsep}{4pt}
    \begin{tabular*}{\textwidth}{@{\extracolsep{\fill}}lccccccc@{}}
        \toprule
        Method & Boxes & Stairs & Mesh Boxes & Inv. high & Wave & \shortstack{Overall\\SR $\uparrow$} & \shortstack{Completion\\$\uparrow$} \\
        \midrule
        Zero & 54.67 & 33.78 & 23.11 & 21.56 & 69.33 & $40.49\pm4.16$ & $74.04\pm1.33$ \\
        Current & 70.44 & 33.56 & 23.78 & 46.22 & \textbf{73.33} & $49.47\pm14.41$ & $78.86\pm8.89$ \\
        Dense & 75.78 & \textbf{46.67} & 27.78 & \textbf{59.11} & 66.44 & $\mathbf{55.16}\pm5.82$ & $\mathbf{82.76}\pm1.53$ \\
        Sparse & \textbf{79.33} & 42.89 & \textbf{30.89} & 40.00 & 69.78 & $52.58\pm10.31$ & $76.78\pm3.29$ \\
        \bottomrule
    \end{tabular*}
\end{table*}


\section{Experiments}
\label{sec:experiments}

We design the experiments to answer four questions:
(1) whether fixed visual histories exhibit terrain- and
motion-dependent trade-offs;
(2) whether \method{} improves locomotion performance
consistently across training seeds and terrain difficulty;
(3) whether temporal-scale learning and state-conditioned
fusion provide complementary benefits; and
(4) whether the resulting policy remains computationally
practical and deployable in simulation and on a physical
robot.

\subsection{Experimental Design}

\paragraph{Shared training settings}
The main comparison is conducted in Isaac Lab using RSL-RL \ac{PPO}. Each configuration is trained for 40,000 iterations with training seeds 42, 43, and 44. At each iteration, 1,024 parallel environments collect 24 rollout steps, followed by three optimization epochs over eight minibatches. The Adam learning
rate is initialized to $7.5\times10^{-5}$ with an adaptive \ac{KL} target of $0.008$. We use $\gamma=0.995$, \ac{GAE} $\lambda=0.97$, a \ac{PPO} clipping ratio of $0.2$, entropy and value coefficients of $0.003$ and $0.5$, respectively, and a
global gradient-norm limit of $0.5$. Dynamics randomization includes friction in $[0.2,1.25]$, additive base mass in $[-5,5]$ kg, link-mass scaling in $[0.9,1.1]$, and perturbations to \ac{PD} gains, observations, and actions. The depth \ac{FIFO} is advanced at the 50-Hz policy rate.

The complex-terrain benchmark comprises ten obstacle classes: normal and high stairs in both directions, discrete and mesh boxes, inverse-pyramid slopes, raised mounds, pits, and wave terrain. Each class contains ten difficulty levels, ranging from level 0 (easiest) to level 9 (most challenging), with configured geometries extending to approximately 23-cm normal steps, 32-cm high steps, 1-m mounds, and 70-cm pit exits. We evaluate 1,000 episodes per terrain--level cell, yielding 100,000 episodes for each trained policy.

\paragraph{Protocol I: Fixed-History Diagnosis}
We compare four matched depth-history configurations: Zero (eight zero-valued channels), Current (the latest supplied depth observation repeated eight times), Dense (eight consecutive \ac{FIFO}
slots), and Sparse (eight slots sampled with stride five). For each of three training seeds, all configurations share the same frozen velocity estimator and visual-neutral initialization, and
are trained for 1,000 \ac{PPO} iterations with 1,024 parallel environments.

Evaluation covers boxes, stairs, mesh boxes, inverse high stairs, and wave terrain at levels $[3,3,3,4,9]$, respectively, under forward commands of $[0.20,0.35,0.50]\,\mathrm{m/s}$. We evaluate 50 paired episodes for each method--seed--terrain--speed combination, yielding 9,000 episodes in total. Depth observations are updated at 30 Hz and
stored in a 50-Hz policy-time \ac{FIFO}. This protocol isolates the effect of fixed history construction and is analyzed separately from the full-curriculum benchmark.

\paragraph{Protocol II: full-curriculum benchmark}
Evaluation uses fixed commands of $v_x=0.35\,\mathrm{m/s}$, $v_y=0$, and $\omega_z=0$, with a 30-s episode limit. Success requires reaching the terrain-specific goal distance while remaining within a $\pm1.0$-m lateral corridor. Goal distances are 3.85 m for forward normal and high stairs, 2.75 m for inverse high stairs, and 3.80 m for all other terrain classes.

We compare Dense, Uniform-Multi, Learned-Multi, Fixed+Gate, Global-Static, and full \method{}. Uniform-Multi uses fixed strides $[0.5,1.5,3.0,4.5]$ with uniform fusion; Learned-Multi learns the strides with uniform fusion; Fixed+Gate combines fixed strides with state-conditioned fusion; and Global-Static learns the strides with observation-independent fusion weights. Full \method{} jointly
learns the temporal strides and performs state-conditioned fusion. All methods share the same locomotion backbone, \ac{PPO} settings, training budget, environment distribution, randomization, estimator schedule, and paired evaluation instances.

\paragraph{Metrics and Statistical Analysis}
\ac{SR} is the percentage of episodes that reach the terrain-specific goal while remaining within the $\pm1.0$-m lateral corridor. Completion is defined as
\begin{equation}
    \mathrm{Completion}
    =
    100\,\operatorname{clip}
    \left(
        \frac{d_{\max}}{d_{\mathrm{goal}}},0,1
    \right),
\end{equation}
where $d_{\max}$ is the maximum forward displacement from the spawn point. Termination rate includes all configured pre-goal contact terminations, not only physical falls. Worst-terrain \ac{SR} is the minimum terrain-aggregated \ac{SR} across the ten obstacle
classes for each seed. We report mean$\pm$\ac{SD} over three training seeds and estimate pairwise 95\% confidence intervals using 10,000 hierarchical-bootstrap resamples. The diagnosis resamples seeds and paired episodes within terrain--speed conditions, whereas the full benchmark resamples seed blocks and paired instances within terrain--level cells with equal cell weighting.
\begin{figure*}[!t]
    \centering
    \includegraphics[width=\textwidth]{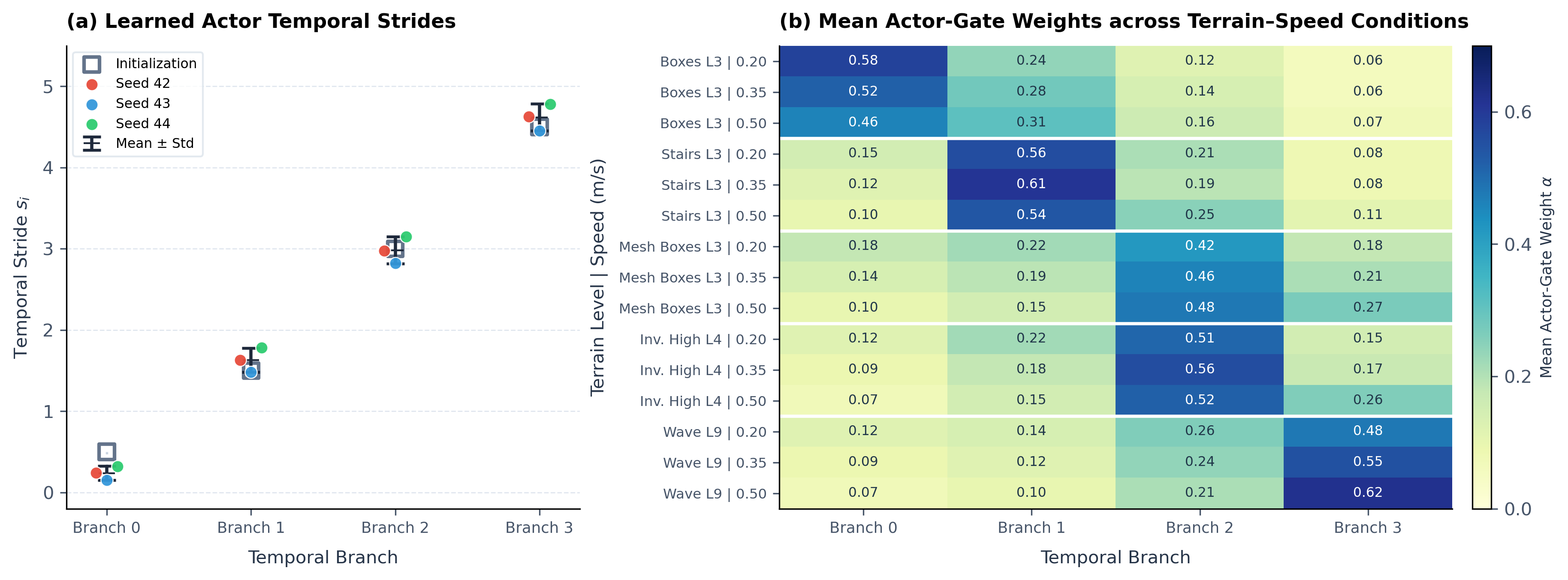}
    \caption{Learned temporal supports and actor-gate usage.
    (a) Final actor strides for seeds 42, 43, and 44, expressed in policy-time \ac{FIFO} slots; hollow squares mark initialization and error bars show mean$\pm$SD across seeds.
    (b) Descriptive mean actor temporal-gate weights for the indicated terrain--speed conditions, obtained by first averaging over policy steps within each episode and then averaging equally over 30 episodes per condition. Branch labels are the original network branch IDs and are not reordered by final learned stride values.}
    \label{fig:temporal_analysis}
\end{figure*}

\begin{table*}[!t]
    \centering
    \caption{Main comparison: 40k \ac{PPO} training iterations, three training seeds, and 100,000 episodes per trained policy. Metrics are percentages (mean$\pm$SD).}
    \label{tab:main_comparison}
    \setlength{\tabcolsep}{5pt}
    \begin{tabular*}{\textwidth}{@{\extracolsep{\fill}}lcccc@{}}
        \toprule
        Method & SR $\uparrow$ & Completion $\uparrow$ & Worst SR $\uparrow$ & Termination $\downarrow$ \\
        \midrule
        Dense & $58.2\pm4.2$ & $65.2\pm3.8$ & $32.1\pm5.6$ & $21.3\pm2.1$ \\
        Uniform-Multi & $62.1\pm3.9$ & $69.0\pm3.5$ & $38.3\pm5.1$ & $18.8\pm1.9$ \\
        Learned-Multi & $68.4\pm3.1$ & $74.6\pm2.9$ & $46.2\pm4.3$ & $14.4\pm1.5$ \\
        Global-Static & $52.3\pm4.8$ & $60.1\pm4.1$ & $25.8\pm6.2$ & $24.6\pm2.4$ \\
        \textbf{\method{}} & $\mathbf{78.4\pm2.3}$ & $\mathbf{83.8\pm1.8}$ & $\mathbf{54.5\pm3.2}$ & $\mathbf{8.2\pm0.9}$ \\
        \bottomrule
    \end{tabular*}
\end{table*}

\subsection{Fixed-History Diagnosis}
\label{sec:fixed_history_results}

Table~\ref{tab:fixed_diagnosis} shows that Dense achieves the
highest aggregate success and completion, but the best fixed
history varies across terrain types. Sparse performs best on
boxes and mesh boxes, Dense performs best on stairs and inverse
high stairs, and Current performs best on wave terrain.

These ranking reversals show that fixed histories provide
different perceptual capabilities rather than forming a
universally ordered set of representations. Dense recent
observations are effective when local geometric evolution is
important, whereas longer or current-only views can be more
useful under other terrain structures and motion conditions.
No single fixed history achieves the highest mean success on
all evaluated terrains, motivating a policy that retains
multiple temporal supports and adapts their utilization online.

\subsection{Full-Curriculum Evaluation and Component Analysis}

\paragraph{Multi-seed performance.}
Table~\ref{tab:main_comparison} reports performance over the
complete ten-level curriculum. \method{} achieves
$78.4\pm2.3\%$ \ac{SR}, compared with $58.2\pm4.2\%$ for Dense,
$62.1\pm3.9\%$ for Uniform-Multi, and $68.4\pm3.1\%$ for
Learned-Multi. The corresponding improvements are 20.2, 16.3,
and 10.0 percentage points, with hierarchical-bootstrap
95\% intervals of $[16.1,24.3]$, $[12.1,19.8]$, and
$[6.4,13.5]$, respectively.

Beyond aggregate success, \method{} achieves a worst-terrain
\ac{SR} of $54.5\pm3.2\%$, compared with $32.1\pm5.6\%$ for Dense.
This indicates that the gain is not confined to already easy
terrain conditions. The \method{}--Dense \ac{SR} difference is
6.2 percentage points at level 0 and 28.5 points at level 9.
This endpoint comparison indicates a larger advantage at the
highest evaluated difficulty, without assuming a monotonic
trend over the intermediate levels.

\paragraph{Temporal-scale learning and state-conditioned fusion.}
Table~\ref{tab:factorial} provides a matched $2\times2$
comparison of temporal-scale construction and fusion. With
fixed temporal scales, replacing uniform fusion with
state-conditioned fusion increases mean \ac{SR} from
$62.1\pm3.9\%$ to $71.5\pm3.5\%$. With learned temporal
scales, the corresponding increase is from $68.4\pm3.1\%$ to
$78.4\pm2.3\%$.

Conversely, learning the temporal scales improves mean \ac{SR} from
$62.1\pm3.9\%$ to $68.4\pm3.1\%$ under uniform fusion and
from $71.5\pm3.5\%$ to $78.4\pm2.3\%$ under
state-conditioned fusion. These matched comparisons show that
temporal-scale learning and state-conditioned utilization
provide complementary benefits, with their combination
achieving the highest mean success.

\paragraph{Dynamic-fusion intervention.}
To test whether the trained policies rely on online fusion
variation, we estimate a mean fusion vector for each
\method{} seed using 5,000 disjoint calibration episodes.
We then replace the observation-dependent fusion weights with
this constant vector while leaving all other parameters
unchanged.

The resulting Frozen-Mean policies achieve
$61.2\pm3.6\%$ \ac{SR}, compared with $78.4\pm2.3\%$ for the
original policies. The dynamic-minus-frozen difference is
17.2 percentage points, with a 95\% hierarchical-bootstrap
interval of $[13.2,21.1]$. Frozen-Mean is a test-time
intervention rather than an independently trained baseline;
the result therefore demonstrates the trained policies'
dependence on online fusion variation, rather than the best
performance attainable by all possible static-fusion methods.

\paragraph{Learned temporal behavior.}
Figure~\ref{fig:temporal_analysis} shows that the learned actor
stride banks retain distinct short- to long-range temporal
supports across all three training seeds. The shortest branch
moves toward denser sampling of recent observations, whereas
the remaining branches retain intermediate and longer
look-back ranges.

The mean gate vectors also vary across terrain--speed
conditions, including across commanded speeds within the same
terrain. These statistics indicate condition-dependent use of
the candidate temporal views. They are descriptive, however,
and do not establish a one-to-one causal specialization between
a particular branch and a particular terrain.

\paragraph{Training stabilization.}
At the reference training seed 42, full \method{} achieves
79.8\% \ac{SR}. Removing the uniform-fusion warm-up, temporal-scale
diversity regularization, or branch-utilization balancing
reduces \ac{SR} to 72.4\%, 58.4\%, and 73.1\%, respectively.
Because these diagnostics use one training seed, they
characterize the evaluated optimization recipe rather than
across-seed robustness.

\begin{table}[!t]
    \centering
    \caption{Factorial ablation of temporal-scale learning and state-conditioned fusion. SR (\%) is reported as mean$\pm$SD over three training seeds under the full-curriculum benchmark.}
    \label{tab:factorial}
    \setlength{\tabcolsep}{4pt}
    \begin{tabular}{lcc}
        \toprule
        \shortstack{Temporal\\scales}
        & \shortstack{Uniform\\fusion}
        & \shortstack{State-conditioned\\fusion} \\
        \midrule
        Fixed
        & $62.1\pm3.9$
        & $71.5\pm3.5$ \\
        Learned
        & $68.4\pm3.1$
        & $\mathbf{78.4\pm2.3}$ \\
        \bottomrule
    \end{tabular}
\end{table}

\subsection{Computational Cost and Deployment}

\paragraph{Computational cost.}
Dense and \method{} contain 1,708,880 and 1,808,352
actor--critic parameters, respectively, excluding the velocity
estimator. Thus, \method{} increases the parameter count by
5.82\%, while the deterministic actor contains 887,648
parameters.

Under a matched RTX 5080 benchmark, the mean and \ac{P95}
actor-forward latencies are 0.304/0.398 ms for Dense and
1.183/1.489 ms for \method{}. Measurements use PyTorch CUDA,
\ac{FP32} inference, batch size one, GPU-resident synthetic inputs,
1,000 warm-up calls, and 10,000 timed calls with CUDA
synchronization after every forward pass. Timing includes
temporal sampling, visual encoding, temporal fusion, and the
locomotion \ac{MoE}, but excludes preprocessing, velocity
estimation, communication, and \ac{PD} control. The measured
1.489-ms \ac{P95} latency is below the 20-ms policy period in this
setting, but should not be interpreted as an end-to-end
worst-case timing guarantee.

\paragraph{Cross-simulator evaluation.}
We deploy the pre-designated seed-42 checkpoint trained for
40,000 iterations, rather than selecting a checkpoint
retrospectively from the three-seed comparison. In MuJoCo, the
policy completes 41 of 45 goal-reaching trials across nine
fixed courses, with five trials per course. Platform ascent
succeeds in 4/5 trials and gap traversal in 2/5 trials; all
remaining courses achieve 5/5. These tests evaluate
cross-simulator execution and are separate from the Isaac Lab
full-curriculum benchmark.

\paragraph{Physical deployment.}
The physical platform is an arm-equipped Unitree B2. The
policy controls the twelve leg joints, while the six arm joints
remain fixed outside the locomotion action space. Depth
observations are transmitted to an external RTX 5080
workstation for policy inference, and Unitree SDK2 manages
robot communication. The operator commands approximately
$0.3\,\mathrm{m/s}$ forward motion.

Across five physical courses and five trials per course,
\method{} completes 23/25 trials. It achieves 5/5 on gravel,
rocky terrain, a sandy slope, and the continuous stair
descent--ascent course, and 3/5 on the pit descent--ascent
course. These experiments demonstrate successful execution of
the complete body-aware temporal-perception and locomotion
pipeline on the tested courses. They do not establish
superiority over untested hardware baselines or unrestricted
terrain reliability.

\begin{table}[!t]
    \centering
    \caption{Physical trials with the seed-42 policy. Dimensions are approximate.}
    \label{tab:real_robot}
    \setlength{\tabcolsep}{3pt}
    \begin{tabular}{lcc}
        \toprule
        Course & Geometry & Completed \\
        \midrule
        Gravel & 1--2~cm particles & 5/5 \\
        Rocky terrain & 8--30~cm rocks & 5/5 \\
        Sandy slope & $\approx$60~cm high & 5/5 \\
        Stairs, down--up & $\approx$10~cm/step & 5/5 \\
        Pit, down--up & $\approx$47~cm deep & 3/5 \\
        \midrule
        \textbf{Total} & -- & \textbf{23/25} \\
        \bottomrule
    \end{tabular}
\end{table}



\section{Conclusion}
\label{sec:conclusion}

In this paper, we introduced \method{}, a body-aware dynamic visual perception framework for complex-terrain quadruped locomotion. \method{} learns complementary temporal receptive fields from a long depth memory and dynamically aggregates their representations using proprioceptive history and
multi-scale terrain evidence. By decoupling global temporal-view discovery from state-conditioned utilization, it enables a single policy to align visual memory with the robot's evolving motion state while preserving the existing locomotion backbone and action interface. Evaluations in simulation and on a physical Unitree B2 demonstrate the effectiveness and practical feasibility of the proposed approach across the tested terrain conditions.

Several limitations remain. The current evaluation uses three training seeds, one locomotion backbone, and a single commanded forward speed in the full-curriculum benchmark. The results therefore establish the benefits of the proposed temporal design under matched settings rather than universal superiority over other temporal encoders or motion conditions. Moreover,
adaptation is restricted to a finite stride bank that remains fixed after training. \ac{FIFO}-index sampling does not explicitly account for physical capture time, repeated camera frames, or ego-motion-induced viewpoint changes, and all four temporal views are encoded at every policy step. The physical evaluation is further limited to one arm-equipped Unitree B2, offboard inference, five trials per course, and no matched hardware
baselines.

Future work will investigate capture-time-aware and motion-compensated visual memory, more efficient temporal routing, state-dependent generation of temporal supports, and broader evaluation across robot morphologies and deployment platforms.

\bibliographystyle{IEEEtran}
\bibliography{references}

\end{document}

%% file: acronyms.tex
\acrodef{PPO}[PPO]{Proximal Policy Optimization}
\acrodef{FIFO}[FIFO]{First-In, First-Out}
\acrodef{GAE}[GAE]{Generalized Advantage Estimation}
\acrodef{KL}[KL]{Kullback--Leibler Divergence}
\acrodef{MoE}[MoE]{Mixture of Experts}
\acrodef{MLP}[MLP]{Multilayer Perceptron}
\acrodef{PD}[PD]{Proportional--Derivative}
\acrodef{SR}[SR]{Success Rate}
\acrodef{SD}[SD]{Standard Deviation}
\acrodef{P95}[P95]{95th Percentile}
\acrodef{FP32}[FP32]{32-bit Floating-Point}
\acrodef{ReLU}[ReLU]{Rectified Linear Unit}
\acrodef{ELU}[ELU]{Exponential Linear Unit}